%% file: main.tex
\documentclass[11pt]{article}

\usepackage[utf8]{inputenc}
\usepackage[T1]{fontenc}
\usepackage{lmodern}
\usepackage[margin=1in]{geometry}
\usepackage{graphics} % for pdf, bitmapped graphics files
\usepackage{subfigure}
\usepackage{epsfig} % for postscript graphics files
\usepackage{amsmath} % assumes amsmath package installed
\usepackage{amssymb}  % assumes amsmath package insta1lled
\usepackage{booktabs}
\usepackage{algorithm}
\usepackage{algpseudocode}
\usepackage{cite}
\usepackage{url}
\usepackage{xcolor}
\usepackage{enumitem}
\usepackage{caption}
\usepackage{multirow}
\usepackage{indentfirst}
\usepackage{IEEEtrantools}

\usepackage[most]{tcolorbox}
\newtcolorbox{caveat}{colback=orange!6,colframe=orange!55!black,boxrule=0.6pt,
  left=6pt,right=6pt,top=4pt,bottom=4pt,arc=2pt}

\begin{document}

\title{\textbf{Large Behavior Model: A Promptable Digital Twin of the Retail Customer}}

\author{
Wachiravit Modecrua, Krittin Pachtrachai, Touchapon Kraisingkorn \\
Amity Research and Application Center (ARAC),\\
Amity AI Holdings Co., Ltd.\\
{\tt\small \{wachiravit, krittin, touchapon\}@amity.co}
}

\maketitle

\bstctlcite{BSTcontrol}

\input{abstract}

\input{introduction}
\input{system}
\input{method}
\input{results}
\input{conclusion}

\bibliographystyle{IEEEtran}
\bibliography{myBib}

\input{appendix}

\end{document}

%% file: abstract.tex
\begin{abstract}
Customer behavior modeling underpins recommendation, marketing, and decision support, yet existing approaches either optimize predictive accuracy without explaining decisions or simulate users without grounding them in real behavioral data. We present the Large Behavioral Model (LBM) that learns customer decision making directly from large-scale retail transactions through a unified Person–Environment formulation. Customer state is represented by a behavioral profile derived from historical purchases, while product context is incorporated through retrieval-augmented generation. The model is trained using continued pre-training on verbalized behavioral data, supervised fine-tuning for decision generation, and reinforcement learning with verifiable rewards for evidence-based calibration.

We evaluate the proposed framework on purchase prediction, hard-negative discrimination, basket completion, promotion response, and cross-domain voucher redemption. The model consistently outperforms frontier general-purpose language models on in-domain retail tasks while demonstrating strong zero-shot and fine-tuned transfer across retailers and decision domains. Ablation studies show that continued pre-training is the primary driver of behavioral generalization, retrieval is most effective when applied during both training and inference, and reinforcement learning improves reliance on explicit behavioral evidence over generic language-model priors. These results demonstrate that behavioral knowledge encoded in transaction histories can be effectively learned by language models, providing a scalable foundation for customer digital twins and behavior simulation.
\end{abstract}

%% file: introduction.tex
\section{Introduction}
\label{sec:intro}

Understanding and predicting human decision-making is a long-standing problem in fields including economics, marketing, recommender systems, and artificial intelligence. Accurate behavioral models enable a wide range of applications, such as personalized recommendation, demand forecasting, promotion planning, pricing optimization, and market research \cite{mcfadden1974conditional,train2009discrete,wedel2016marketing}. Despite decades of research, faithfully modeling the behavior of an individual remains challenging because decisions arise from the interaction between relatively stable personal preferences and the context in which choices are made. This perspective is consistent with Lewin's field theory, which characterizes behavior as a function of both the person and the environment \cite{lewin1936principles}. Consequently, models that capture only population-level patterns often fail to explain why two seemingly similar customers make different decisions under the same circumstances.

In retail, this challenge is particularly evident. Large retailers routinely make pricing, promotion, and assortment decisions for millions of customers, yet their understanding of customer preferences is still largely informed by surveys, focus groups, and consumer panels \cite{malhotra2019marketing,wedel2016marketing}. Although these methods remain indispensable in marketing research, they are expensive, slow to update, and typically cover only a small fraction of the customer population. More importantly, stated preferences collected through questionnaires often differ from revealed preferences observed through actual purchasing behavior, motivating the increasing use of transaction data for behavioral modeling \cite{guadagni1983logit}. Consequently, retailers possess abundant behavioral data but relatively limited tools for transforming those observations into faithful simulations of individual customer decisions.

To address this limitation, previous research has developed a wide range of behavioral models. Classical discrete choice models estimate purchasing probabilities from utility-maximization principles \cite{mcfadden1974conditional,train2009discrete}, while recommender systems and sequential recommendation models learn user representations from historical interactions to predict future purchases \cite{koren2009matrix,he2017neural,kang2018sasrec,sun2019bert4rec,hstu,tiger}. These approaches have achieved considerable success in tasks such as purchase prediction, recommendation, and sequential decision modeling. However, they are generally optimized for a single downstream task, requiring separate models for recommendation, promotion response, basket completion, demand forecasting, and related applications. As a result, they do not provide a unified behavioral model capable of answering diverse questions about the same customer.

Recent advances in large language models (LLMs) suggest a different direction. Modern foundation models exhibit strong reasoning, instruction-following, and in-context learning capabilities, allowing a single model to perform a wide variety of tasks through natural-language prompting alone \cite{ouyang2022instructgpt,dubey2024llama3,qwen3technicalreport,gemma2}. This flexibility has inspired a growing body of research on using LLMs to simulate human behavior, including social interactions, survey responses, economic decision making, and online shopping behaviors \cite{park2023generative,argyle2023outofone,stanford1000,lcbm,opera}. More recent work has explored personalized shopping agents through reinforcement learning and behavioral datasets \cite{shopr1,customerr1}. Despite these promising developments, existing approaches predominantly represent individuals using demographic information, manually designed personas, or synthetic profiles \cite{scope,nemotron,survey}. While these representations capture broad population characteristics, they contain limited evidence of an individual's actual behavioral history. Consequently, simulated responses often reflect the prior knowledge encoded in the foundation model rather than the behavioral patterns exhibited by the individual being simulated, leading to a persistent gap between simulated and real-world behavior \cite{sim2real,funhouse,longcontext}.

Recent work has also explored personalizing language models through user representations and embeddings to better capture user-specific preferences \cite{userllm,collabemb,curp}. However, these approaches focus primarily on improving recommendation or personalization rather than faithfully simulating individual decision-making across diverse behavioral tasks.

This paper investigates a simple but fundamental question: \emph{Can language models become faithful simulators of individual consumer behavior when grounded in longitudinal behavioral evidence rather than generic language-model priors?}

We propose the \textbf{Large Behavior Model (LBM)}, a language-model framework for personalized behavioral simulation that grounds customer decisions in two complementary sources of information: persistent behavioral evidence derived from historical transactions and the current decision environment. Inspired by Lewin's formulation that behavior emerges from the interaction between the person and the environment \cite{lewin1936principles}, we model customer behavior as

\begin{equation}
B=f(P,E),
\label{eq:behavior}
\end{equation}

where $P$ denotes a customer's persistent behavioral representation learned from longitudinal purchasing history, and $E$ represents the products and contextual information available at the time of decision. In LBM, the behavioral representation is learned through lightweight personalization modules, while contextual information is supplied dynamically through retrieval-augmented generation (RAG) \cite{lewis2020rag}. This explicit separation allows the same underlying language model to simulate a broad range of customer behaviors—including purchase prediction, basket completion, promotion response, market survey answering, and review generation—without training separate models for each task.

Extensive experiments on large-scale retail transaction data demonstrate that grounding language models in longitudinal behavioral evidence substantially improves customer-level decision simulation. Across multiple behavioral tasks, LBM consistently outperforms frontier foundation models while generalizing across retailers, product catalogs, and application scenarios. These results suggest that behavioral evidence provides a stronger foundation for personalized decision simulation than relying solely on the generic priors encoded in large language models. Moreover, because personalization is achieved through lightweight behavioral adaptation rather than separate models for individual customers, the proposed framework scales efficiently to millions of users while remaining practical for self-hosted deployment. The contributions of this work are summarized as follows.

\subsection*{Contributions}

\begin{enumerate}[leftmargin=1.4em,itemsep=2pt,topsep=2pt]

\item \textbf{A formulation for personalized behavioral simulation.}
We formulate customer behavior as the interaction between persistent behavioral characteristics and the decision environment, extending classical behavioral theory to language-model-based simulation and providing a unified framework for modeling diverse customer behaviors.

\item \textbf{The Large Behavior Model (LBM).}
We propose a personalized language-model architecture that grounds behavioral simulation in longitudinal transaction histories through behavioral representations and retrieval-augmented contextual grounding. Unlike conventional recommender systems \cite{koren2009matrix,he2017neural,kang2018sasrec,hstu} that solve individual prediction tasks, LBM supports multiple behavioral simulation tasks using a single shared language model.

\item \textbf{A scalable behavioral learning framework.}
We develop a four-stage training pipeline combining continual pre-training, supervised fine-tuning, and reinforcement learning to align language models with customer behavior. Comprehensive ablation studies quantify the contribution of each stage and highlight the importance of jointly learning behavioral representations and contextual grounding.

\item \textbf{A benchmark for customer-level behavioral simulation.}
We construct a benchmark spanning multiple retail decision-making tasks—including purchase prediction, basket completion, promotion response, and survey simulation—with carefully designed hard negatives and varying levels of behavioral ambiguity to evaluate customer-specific reasoning.

\item \textbf{Evidence that behavioral grounding outperforms foundation-model priors.}
Across both in-domain and cross-domain evaluations, we demonstrate that grounding language models in longitudinal behavioral evidence consistently improves personalized decision simulation over frontier foundation models. These findings suggest that behavioral evidence is a more effective basis for simulating individual consumer behavior than relying solely on the generic priors encoded in large language models.

\end{enumerate}

%% file: system.tex
\section{System Overview}

We present the Large Behavior Model (LBM), a framework for simulating individual consumer decisions from historical transaction data. Rather than learning a separate model for each downstream task, LBM models customer behavior as the interaction between a persistent behavioral profile and a dynamic decision environment. This decomposition enables a single language model to generalize across multiple retail decision tasks—including purchase prediction, basket completion, promotion response, and survey-style preference elicitation—using only prompt conditioning.

\begin{figure}[h]
\centering
\includegraphics[width=\linewidth]{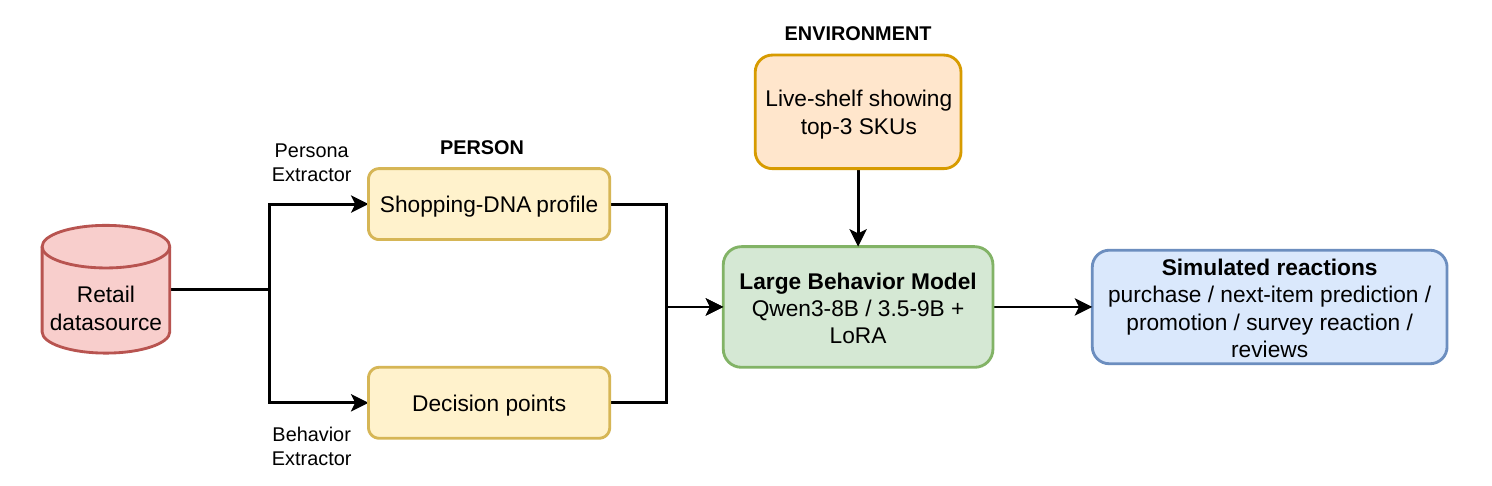}
\caption{From raw transactions to simulated decisions. The Persona Extractor summarizes stable demographic/lifestage traits while a Behavior Extractor derives decision points from raw transactions, together forming the Shopping-DNA profile. Person is supplied by a per-segment LoRA adapter plus a system prompt built from this profile; Environment is supplied by RAG at decision time. The same weights answer purchase, basket, promotion, survey, and review questions, selected by the prompt.}
\label{fig:overview}
\end{figure}

Figure~\ref{fig:overview} provides an overview of the complete system. Raw retail transactions are transformed into behavioral representations and supervised decision examples, while a retrieval module injects product knowledge at inference time to construct the decision environment.

\paragraph{Overview of the pipeline.}

The pipeline consists of three stages. First, historical transaction logs are processed to construct a persistent behavioral representation for each customer. Second, real purchase events are converted into supervised behavioral decision examples spanning multiple prediction tasks. Finally, during inference, the model combines the customer representation with retrieved product knowledge to simulate decisions under the current shopping context. This design separates long-term customer behavior from short-term environmental context while allowing both to be encoded in natural language.

\paragraph{Person--Environment decomposition.}

We hypothesize that customer decisions arise from the interaction of persistent behavioral preferences and the current decision context. We therefore formulate behavioral simulation as modeling two complementary factors: Person and Environment.

\begin{itemize}
    \item \textbf{Person ($P$):} a persistent representation derived from historical purchasing behavior that captures long-term preferences, shopping habits, and behavioral tendencies.
    \item \textbf{Environment ($E$):} a dynamic representation of the products available in the current decision context, retrieved from the product knowledge base.
\end{itemize}

Consumer behavior is therefore modeled as

\begin{equation}
B = f(P,E),
\end{equation}

where the language model predicts decisions conditioned jointly on the customer representation and the surrounding shopping environment.

This separation is central to our framework. Rather than memorizing customer-specific behavior within model parameters, customer identity is supplied through prompt conditioning, while environmental knowledge is retrieved dynamically at inference time. The resulting architecture naturally generalizes to unseen customers without retraining. The design also enables the same underlying model to be reused across multiple behavioral tasks—including purchase prediction, basket completion, promotion response, and survey-style reasoning—without requiring task-specific architectures.

\subsection{Data}
\label{sec:data}

We construct our dataset from large-scale retail transaction logs and product catalog information. The data supports two core purposes: (i) learning persistent behavioral representations of customers, and (ii) constructing realistic decision environments for behavioral simulation.

\subsubsection{Transaction data}

The core dataset consists of anonymized retail transaction records, where each entry corresponds to a single product purchased within a customer basket. Each record contains a masked customer identifier, timestamp, product identifier, quantity, unit price, applied discounts, and final net spend.

Transactions are aggregated into baskets, representing individual shopping trips. Each basket contains multiple items, forming sequential decision contexts that reflect real-world purchasing behavior. Products are structured within a hierarchical taxonomy spanning division, section, and item levels, enabling multi-resolution behavioral modeling.

From these logs, we derive:
(i) historical purchase sequences per customer, used to construct behavioral representations, and
(ii) observed purchase decisions, used as supervision signals for learning decision-making behavior.

\subsubsection{Product catalog}

We construct a product knowledge base containing all available SKUs, along with structured metadata including product names, category hierarchy, and price statistics.

Each product is embedded into a dense vector space to enable semantic retrieval of similar items. This retrieval mechanism is used to construct the decision environment at inference time by selecting a small set of relevant candidate products.

\subsubsection{Dataset scale}

\begin{table}[h]
\centering
\small
\caption{Dataset summary for the primary retail cohort.}
\label{tab:data}
\begin{tabular}{lrr}
\toprule
 & \textbf{Train} & \textbf{Test} \\
\midrule
Transactions          & 159{,}667 & 40{,}665 \\
Customers             & 1{,}500   & 1{,}500 (held-out baskets) \\
Products (catalog)    & 38{,}516   & (same) \\
\bottomrule
\end{tabular}
\end{table}

In addition to the primary cohort, we construct multiple evaluation splits for behavioral simulation, including held-out within-domain test sets and harder negative-sampling-based evaluations designed to test decision boundaries under ambiguity. We further evaluate cross-domain generalization on external retail and consumer behavior datasets. Table~\ref{tab:data} summarizes the primary dataset used in this work.

\subsubsection{Summary of usage}

The dataset serves two roles in our framework:
(i) constructing persistent customer behavioral representations, and
(ii) defining realistic decision environments for simulation at inference time.

%% file: method.tex
\section{Method}
\label{sec:method}

The Large Behavior Model (LBM) treats customer decision making as a conditional reasoning problem. Rather than learning a separate model for each downstream task, the model conditions on two complementary sources of information: (i) a persistent representation of the customer derived from historical transactions, and (ii) a dynamic decision environment describing the products currently under consideration. This factorization enables a single language model to simulate multiple retail behaviors while sharing a common behavioral representation.

\begin{figure}[h] \centering \includegraphics[width=0.8\linewidth]{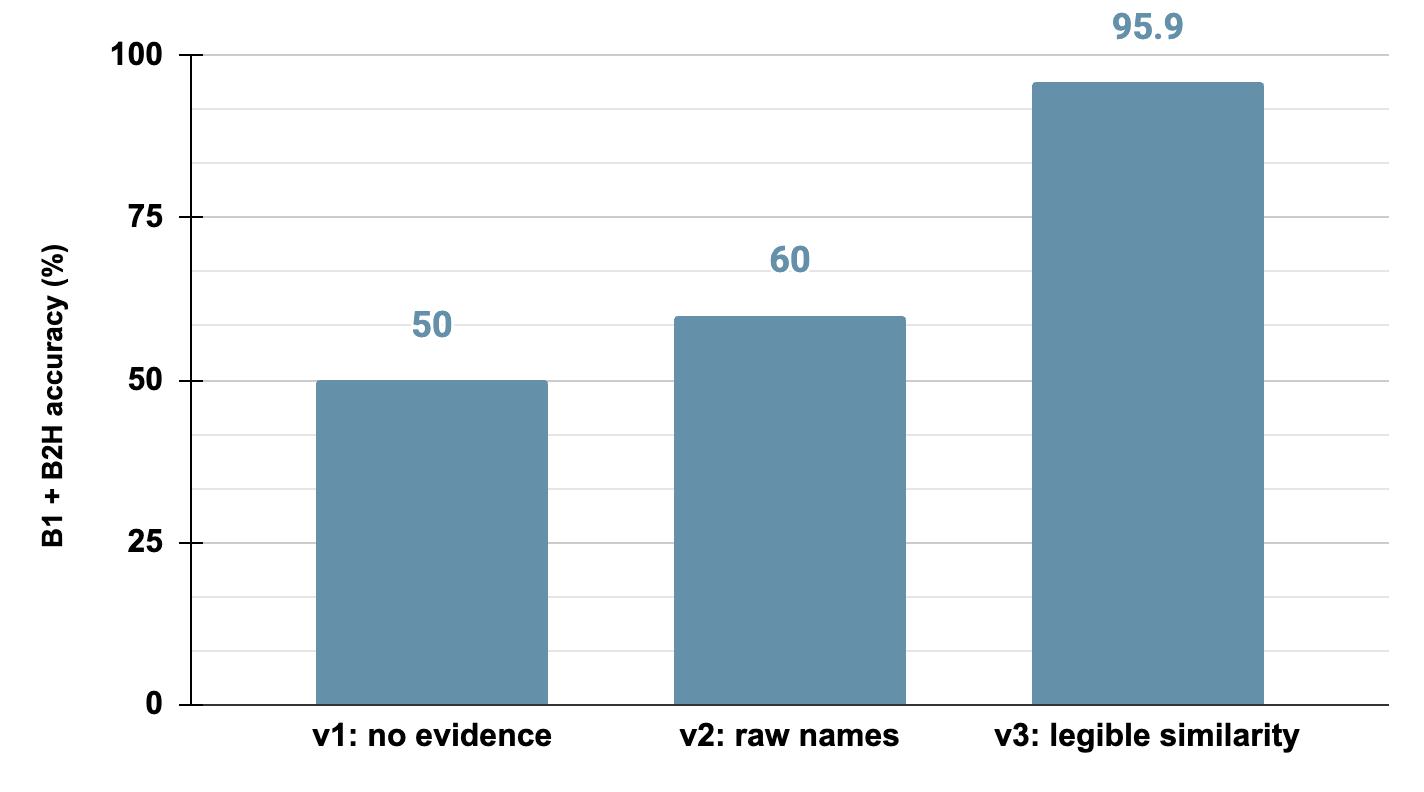}
\caption{The B-hard difficulty ladder: the model represents a customer only to the resolution the prompt
encodes. With no in-prompt evidence (v1) accuracy is near chance; raw similar-item names (v2) reach
$\sim$60\%; a single \emph{legible} similarity line (v3) reaches 95.9\% (vs.\ 73.4\% for GPT-5.5). The
bottleneck is prompt signal, not weights.} \label{fig:b-hard}
\end{figure}

Figure~\ref{fig:pipeline} summarizes the complete training pipeline, while Figure~\ref{fig:b-hard} illustrates how progressively richer behavioral evidence affects downstream decision quality.

\subsection{Problem formulation}

We formulate behavioral prediction as

\begin{equation}
B = f(P,E),
\end{equation}

where $P$ denotes the persistent customer representation and $E$ denotes the decision environment available at inference time.

The persistent representation captures long-term purchasing tendencies, whereas the environment describes the immediate choice context. This decomposition allows the same underlying model to support purchase prediction, basket completion, promotion response, survey reasoning, and other consumer decision tasks without task-specific architectures.

\subsection{Customer representation}
\label{sec:customer-rep}
Each customer is represented using two complementary components: a structured behavioral profile and a lightweight behavioral adapter.

\paragraph{Shopping-DNA profile.}

We first construct a structured prompt exclusively from transactions observed during the training period. The resulting Shopping-DNA summarizes stable behavioral characteristics including demographic information (when available), life stage, price sensitivity, category preferences, repeat purchasing patterns, basket statistics, temporal shopping behavior, channel usage, and promotion affinity. No information from the evaluation period is included, preventing future information leakage.

\paragraph{Behavioral adapters.}

To complement the prompt representation, we fine-tune LoRA adapters on behavioral decision data.

Two adaptation strategies are investigated.

\begin{itemize}
\item \textbf{Segment-level adapters} are trained on customer groups defined by life stage or retail personas. These adapters provide complete customer coverage, including cold-start users.
\item \textbf{User-level adapters} are trained separately for individual customers. Although they can outperform segment models when abundant history is available (typically 80--300 behavioral examples), they do not generalize to sparse or unseen customers.
\end{itemize}

Consequently, all experiments in this paper adopt segment-level adaptation as the primary deployment strategy.

\subsection{Decision environment via retrieval}

The second component of the framework models the customer's immediate decision environment.

Each product in the retail catalog is embedded into a dense vector space, and a ChromaDB index is used to retrieve the top-$k$ semantically similar products (typically $k=3$) using cosine similarity. These retrieved products are injected directly into the model prompt during behavioral reasoning.

Our ablation studies demonstrate that retrieval is effective only when it remains part of the decision context. Incorporating retrieved information during both supervised training and inference substantially improves prediction accuracy, whereas inserting retrieval into the continual pre-training corpus degrades performance. These results indicate that retrieval functions as contextual evidence rather than model knowledge.

\subsection{Four-stage training pipeline}

\begin{figure}[h] \centering 
\includegraphics[width=\linewidth]{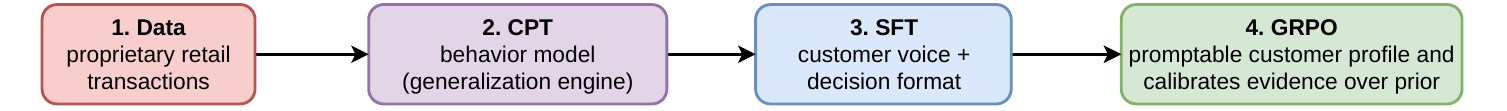}
\caption{A prompt change re-instantiates any customer; CPT generalizes, SFT format-overfits, GRPO calibrates evidence-over-prior trust. The four-stage recipe chained on \emph{verbalized} behavior. Each stage adds capability a zero-shot frontier model lacks; CPT is the backbone, SFT supplies the decision interface, and GRPO generalizes and calibrates trust in the in-prompt evidence.} \label{fig:pipeline}
\end{figure}

The LBM is trained using a progressive four-stage pipeline illustrated in Figure~\ref{fig:pipeline}. Rather than optimizing a single objective, each stage contributes a distinct capability.

\begin{enumerate}
\item \textbf{Behavioral data collection} provides large-scale retail transaction histories.
\item \textbf{Continual pre-training (CPT)} adapts the language model into a behavior model capable of learning general purchasing patterns.
\item \textbf{Supervised fine-tuning (SFT)} teaches structured decision formats and behavioral reasoning over transaction-derived prompts.
\item \textbf{GRPO reinforcement learning} calibrates decision making by encouraging reliance on explicit behavioral evidence while discouraging generic language-model priors.
\end{enumerate}

Figure~\ref{fig:pipeline} illustrates how these stages progressively transform a general-purpose language model into a customer behavior simulator.

Our empirical study consistently shows that CPT provides the majority of transferable behavioral knowledge, while SFT primarily improves response formatting. GRPO further strengthens robustness by calibrating the model's confidence according to evidence contained in the prompt.

\subsection{B-task formulation}

We evaluate the model on four behavioral prediction tasks constructed from real transaction logs.

\begin{table}[h]
\centering
\small
\caption{Behavioral evaluation tasks (B1--B4).}
\label{tab:btasks}
\begin{tabular}{p{1.0cm}p{3.6cm}p{8.4cm}}
\toprule
\textbf{Task} & \textbf{Objective} & \textbf{Construction} \\
\midrule
B1 & Purchase (positive) & Actual purchased items with retrieved context. \\
B2 & Purchase (negative) & Hard negatives sampled from same and different sections. \\
B3 & Basket completion & Predict next item given partial basket context. \\
B4 & Promotion response & Evaluate response to discounts vs counterfactual offers. \\
\bottomrule
\end{tabular}
\end{table}

To construct robust decision boundaries, negative samples are generated using both intra-category and cross-category perturbations.

\subsection{B-hard evaluation and prompt sensitivity}

We further define a difficulty ladder (B-hard) to evaluate sensitivity to contextual signal strength. Performance increases sharply when explicit similarity information is included in the prompt, demonstrating that model performance is strongly driven by the quality of in-context evidence rather than internal computation of similarity.

This indicates that the model primarily operates as an evidence-conditioned decision system rather than a latent similarity engine.

\subsection{Training dynamics and key findings}

We summarize four main empirical findings:

\begin{itemize}
    \item \textbf{CPT provides generalization, SFT improves format adherence.} CPT is responsible for transfer performance, while SFT primarily shapes output structure.
    \item \textbf{Retrieval must be present at inference time.} Removing retrieval significantly reduces performance, while misplacing it in pre-training is detrimental.
    \item \textbf{Reasoning supervision is task-dependent.} Rationale training improves transfer tasks but may reduce performance on fixed-format evaluation.
    \item \textbf{Data packing affects learning stability.} Improper multi-turn packing leads to shortcut learning and degraded generalization.
\end{itemize}

\subsection{Key insight: evidence over prior}

A central property of the model is that decisions are primarily driven by explicit behavioral evidence in the prompt rather than generic priors learned during pre-training. This is reinforced through reinforcement learning, which calibrates the model’s reliance on observed customer history over population-level stereotypes.

%% file: results.tex
\section{Experiments and Results}
\label{sec:results}

\subsection{In-domain Evaluation: Real Retailer}

We evaluate the LBM on held-out transaction decision tasks (Table~\ref{tab:classicb}) and a harder negative-sampling variant (Table~\ref{tab:bhard}). On the classic-B benchmark ($n=5{,}091$), the LBM outperforms GPT-5.5 on average and across all tasks except a near tie on B1. On the harder B-hard v4 set ($n=2{,}490$), which introduces adversarial hard negatives, the performance gap increases to +15.7pp on average, driven primarily by large gains on B2.

\begin{table}[h]
\centering
\small
\caption{Classic-B held-out evaluation ($n=5{,}091$). Higher is better.}
\label{tab:classicb}
\begin{tabular}{lccccc}
\toprule
\textbf{Model} & B1 & B2 & B3 & B4 & \textbf{AVG} \\
\midrule
Our LBM (base + SFT-Retailer-data) & 84.2 & \textbf{97.6} & \textbf{71.1} & \textbf{76.0} & \textbf{82.2} \\
GPT-5.5 & \textbf{84.8} & 75.5 & 57.9 & 72.8 & 72.7 \\
\bottomrule
\end{tabular}
\end{table}

\begin{table}[h]
\centering
\small
\caption{B-hard v4 with hard negatives ($n=2{,}490$). Higher is better.}
\label{tab:bhard}
\begin{tabular}{lccccc}
\toprule
\textbf{Model} & B1 & B2 & B3 & B4 & \textbf{AVG} \\
\midrule
Our LBM (GRPO v5 + SFT-Lazada) & \textbf{98.6} & \textbf{98.9} & \textbf{81.6} & \textbf{52.0} & \textbf{82.8} \\
GPT-5.5 & 94.5 & 51.4 & 70.9 & 51.6 & 67.1 \\
\bottomrule
\end{tabular}
\end{table}

Across both benchmarks, gains are most pronounced on hard-negative discrimination (B2), indicating that performance improvements are driven by improved evidence grounding rather than memorization.

\begin{figure}[h]
\centering
\includegraphics[width=0.8\linewidth]{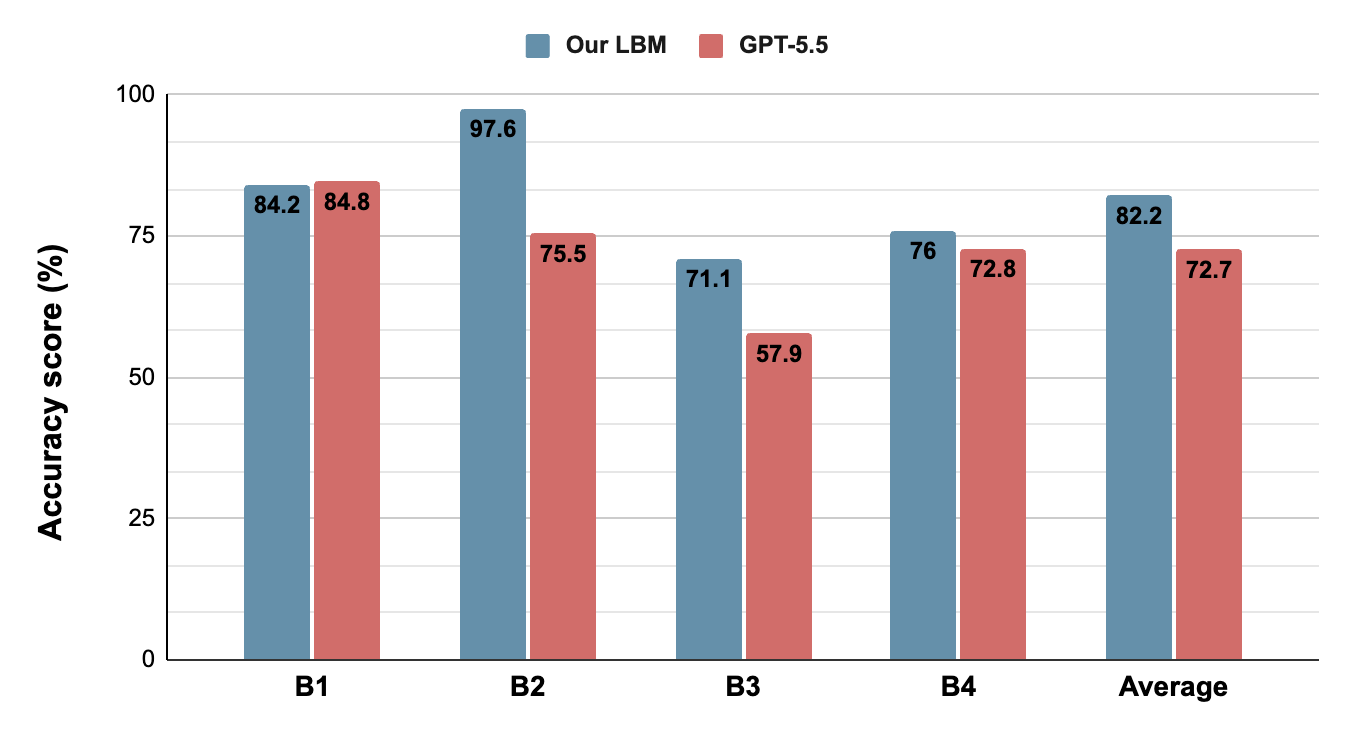}
\caption{In-domain task-by-task comparison on classic-B (held-out, $n = 5,091$). The LBM matches the frontier on B1 and beats it on hard negatives (B2), basket (B3), promotion (B4), and on average (+9.5 points).}
\label{fig:indomain}
\end{figure}

\subsubsection{Stage-wise scaling results}

We further evaluate system performance across deployment scales. In a proof-of-concept setting (10 users), Segment-LoRA + RAG achieves 78.0\% accuracy on purchase tasks (B1+B2), compared to 75.9\% for GPT-5.5 and 72.5\% for the base model. At larger scale (662 users, 24,979 test prompts), the LoRA lift increases substantially, reaching +15.2pp for purchase prediction, while promotion prediction shows weaker gains (+2.7pp), indicating reduced generalization under weaker signal conditions.

Overall, the best configuration (v5 + SFT) achieves 88.7\% average accuracy across B1–B4, compared to 80.1\% for GPT-5.5, corresponding to consistent gains of +9–13pp across tasks.

\subsubsection{GRPO-based calibration}

We train the model through a reinforcement learning curriculum:
SFT-v5 $\rightarrow$ GRPO-v1 (300 steps; verifiable YES/NO rewards; NO-class upweighted; Jaccard reward for B3) $\rightarrow$ GRPO-v4/v5 (adding B4 affinity and B3 co-occurrence constraints).

On the D2 trip-discrimination task (2AFC real vs decoy), performance improves from 54.6\% (SFT-v5) to 68.9\% (GRPO-v5 + SFT), compared to 81.2\% for GPT-5.5. The remaining gap reflects residual limitations in reward calibration rather than representation capacity. This gap is consistent with the segment-level (rather than per-user) adapter granularity used here: D2's decoy-discrimination task may require finer-grained purchase-sequence signal than the current segment adapters encode, an avenue we leave to future user-level adapter scaling (Section \ref{sec:customer-rep}).

\subsection{Out-of-domain Evaluation: Lazada Voucher Redemption}

We evaluate cross-domain generalization on the DMBGN benchmark (SIGKDD’21), consisting of 12,480 held-out cases with no session overlap with training data. The task requires ranking voucher redemption likelihood, measured via AUC.

This setting introduces three simultaneous shifts: (i) domain shift (grocery $\rightarrow$ e-commerce), (ii) geography shift (Thailand $\rightarrow$ SEA-wide), and (iii) task shift (purchase prediction $\rightarrow$ voucher redemption).

\begin{figure}[h]
\centering
\includegraphics[width=0.8\linewidth]{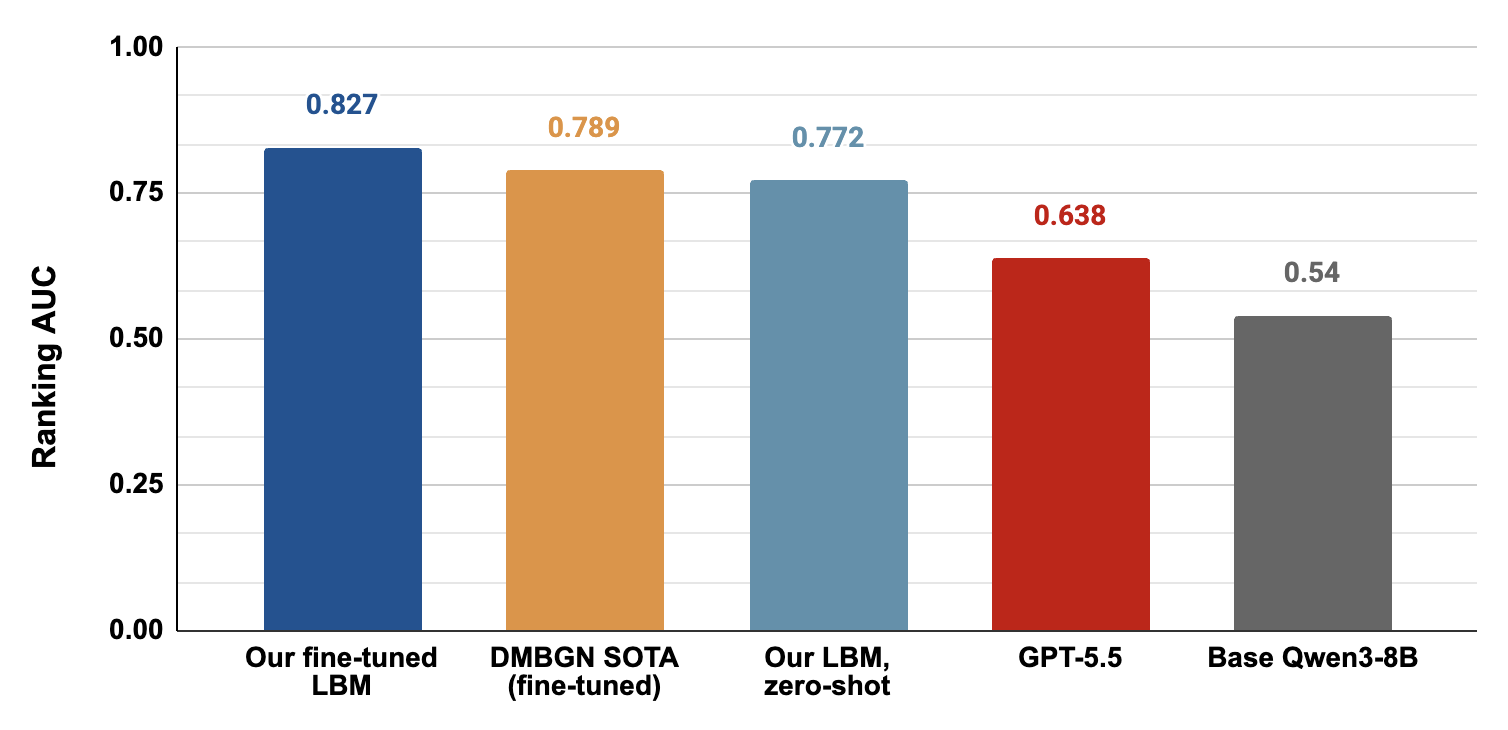}
\caption{Voucher-redemption ranking accuracy (AUC) on the DMBGN benchmark (SIGKDD’21). Trained only on Retailer Data, the LBM transfers zero-shot to Lazada (0.772, above the frontier GPT-5.5); fine-tuned on
Lazada it reaches 0.827. Both LBM bars (blue) clear the GPT-5.5 baseline at every level.}
\label{tab:lazada}
\end{figure}

Despite being trained only on Retailer data, the model achieves strong zero-shot transfer (0.772 AUC), surpassing GPT-5.5. After fine-tuning on Lazada data, performance increases to 0.827 AUC, outperforming both GPT-5.5 and prior SOTA baselines.

Notably, the model remains competitive even in the zero-shot regime, suggesting that behavioral representations learned from retail transactions generalize across platforms when prompt structure is preserved.

\subsubsection{Cross-market transfer analysis}

We further evaluate transfer across heterogeneous datasets:

\textbf{Online Shoppers (UCI, Turkey):} The base model achieves 0.849 AUC (log-prob), close to a strong GBDT ceiling (0.933). SFT reduces performance (0.781), consistent with format overfitting.

\textbf{Tmall (IJCAI-15):} Zero-shot prompt performance ranges from 0.53–0.59 AUC, improving to 0.619 with enriched prompts, but remaining below feature-engineered SOTA (0.703), indicating loss of structured temporal signals.

\textbf{Shopee (SEA):} GPT-5.5 dominates (0.825 vs ~0.65), suggesting that when signal is embedded in platform-specific identifiers or engineered features, prompt-based generalization is insufficient.

Overall, we observe a consistent pattern: the model transfers effectively when behavioral signals are expressible in natural language or coarse structured features, but degrades when predictive signal is encoded in IDs or high-dimensional engineered features.

\subsubsection{Cross-dataset survey transfer (Twin-2K-500)}

We evaluate a reasoning-chain LoRA trained on Amazon grocery survey data on 414 unseen users (16,560 items). The model achieves 71.7\% accuracy, outperforming GPT-5.5 (70.6\%) and a base aggregation baseline (70.3\%).

Key findings are: (i) cross-person augmentation improves generalization, (ii) class balance must match deployment distributions, (iii) reasoning supervision improves structural consistency rather than memorized answers, and (iv) prompt-level aggregation outperforms baked-in aggregation mechanisms.

\section{Discussion}
\label{sec:related}

\paragraph{Behavior simulation is an information problem.}
A central finding of this work is that behavioral simulation is primarily an information problem rather than a model-size problem. Across our experiments, the largest performance gains arise from improving how customer behavior is represented---through Shopping DNA, retrieval-augmented decision contexts, and prompt construction---rather than simply scaling the underlying language model. In particular, the B-hard experiments demonstrate that providing explicit behavioral evidence dramatically improves decision quality, while stronger frontier models without such evidence remain limited by generic behavioral priors. These results suggest that faithful customer representation is a more important bottleneck than model capacity for behavioral simulation.

\paragraph{Evaluation scope and future directions.}

Our evaluation focuses on behavioral decision-making tasks with objective ground truth, including purchase prediction, basket completion, promotion response, and cross-domain transfer. These tasks permit quantitative evaluation against real customer decisions and standardized benchmarks.

Although the LBM can also be prompted to generate customer-facing natural language---such as product reviews, shopping rationales, or survey responses---these capabilities are not quantitatively evaluated in the current work. Unlike purchase decisions, there is generally no unique ground-truth review for a given customer and product, making automatic evaluation substantially more challenging. Consequently, review generation should be viewed as a qualitative capability enabled by the same behavioral representation rather than as a validated benchmark in this paper.

An important direction for future work is the development of evaluation protocols for open-ended behavioral generation, including review synthesis, conversational shopping agents, and virtual consumer studies. Such evaluations will likely require human assessment or preference-based metrics in addition to traditional automatic benchmarks, complementing the decision-centric evaluation presented here.

\paragraph{Behavioral language models.}
Continued pre-training on verbalized behavioral data has been shown to yield strong customer simulation models (e.g., Adobe LCBM~\cite{lcbm}). SFT further improves alignment with behavioral outputs (OPeRA~\cite{opera}). Reinforcement learning with verifiable rewards has recently emerged as a scalable approach to behavioral modeling (Customer-R1~\cite{customerr1}, Shop-R1~\cite{shopr1}), often incorporating persona conditioning and difficulty-aware reward shaping.

Our work extends this line by explicitly chaining CPT, SFT, and GRPO on a unified verbalized behavioral representation, with a clean separation between \textit{person} and \textit{environment} components.

We contrast this with token-level recommender systems (e.g., HSTU~\cite{hstu}, TIGER~\cite{tiger}), which scale effectively for ranking but do not support simulation or language-conditioned reasoning.

\paragraph{Prompt conditioning and generalization.}
Recent studies demonstrate that prompt-based persona conditioning can match or exceed per-user fine-tuning in many settings (Stanford 1,000-person study~\cite{stanford1000}; Twin-2K-500~\cite{twin2k}). Demographic-only representations are weak predictors of behavior (SCOPE~\cite{scope}), while richer behavioral summaries significantly improve fidelity. Synthetic persona corpora (Nemotron~\cite{nemotron}) provide complementary prior distributions.

\paragraph{Fidelity limits of simulation.}
Prior work consistently shows a ceiling in behavioral fidelity. Full persona models often collapse toward demographic baselines (Funhouse Mirrors~\cite{funhouse}), and long-context conditioning can degrade performance beyond a threshold (Long Context, Less Focus~\cite{longcontext}). These findings align with our observation that prompt signal quality, rather than model capacity, is the limiting factor.

\paragraph{Evaluation methodology.}
Reliable evaluation requires (i) real held-out decisions, (ii) hard negative sampling, and (iii) variance-aware metrics rather than point accuracy alone. Prior work (CMU Sim2Real~\cite{sim2real}) shows that LLM-based simulators can overestimate downstream performance due to excessive cooperativeness and reduced behavioral friction. Our evaluation explicitly avoids model-generated users and uses real transaction data with adversarial negatives.

We further highlight that spurious artifacts (e.g., price-ending leakage in prior datasets) can inflate performance; our evaluation explicitly controls for such effects.

%% file: conclusion.tex
\section{Conclusion and Future Work}
\label{sec:conclusion}

We presented a behavioral language modeling framework for customer simulation that combines continued pre-training, supervised fine-tuning, reinforcement learning, and retrieval-augmented prompting within a unified Person--Environment formulation. Rather than treating customer modeling as a recommendation problem over tokenized interaction histories, our approach models purchasing behavior as language-conditioned decision making, allowing the same model to simulate, explain, and predict customer actions.

Across multiple retail benchmarks, the proposed framework consistently outperformed frontier general-purpose language models on behavior prediction tasks. In the in-domain evaluation, the model achieved substantial improvements on purchase prediction, hard-negative discrimination, basket completion, and promotion response while remaining competitive on simpler purchase decisions. More importantly, the learned behavioral representation generalized beyond the training domain, transferring from grocery retail to e-commerce voucher redemption with strong zero-shot performance and further improvements after task-specific fine-tuning. These results suggest that behavioral knowledge acquired from transaction histories can transfer across retailers, markets, and decision types when customer state and environmental context are represented explicitly.

Our experiments also provide several empirical insights into behavioral language modeling. Continued pre-training serves as the primary source of behavioral generalization, whereas supervised fine-tuning mainly teaches the desired response format. Retrieval is most effective when incorporated during both training and inference, but not during continued pre-training. Finally, reinforcement learning improves the model's calibration by encouraging decisions to rely on explicit behavioral evidence rather than generic language-model priors.

Despite these encouraging results, several limitations remain. The current approach still relies on handcrafted textual summaries of customer behavior, which inevitably compress rich transactional histories into natural language descriptions. Performance also degrades when predictive signals are primarily encoded in platform-specific identifiers or highly engineered numerical features rather than semantically interpretable behavioral patterns. Furthermore, while prompt conditioning enables strong zero-shot personalization, it does not learn continuous user representations that can evolve over time.

Future work will focus on three directions. First, we plan to replace manually constructed behavioral summaries with learned customer representations that map transaction histories directly into continuous prompt embeddings or soft prompts. Second, we will extend the framework to larger multi-region and multi-vertical behavioral datasets to improve robustness and cross-domain transfer. Finally, we aim to explore downstream applications beyond purchase prediction, including personalized recommendation, counterfactual customer simulation, virtual market experimentation, and decision-support systems. More broadly, we believe behavioral language models provide a promising foundation for scalable digital twins that bridge structured behavioral data with the reasoning capabilities of modern language models.

%% file: appendix.tex
\appendix
\section{Reproducibility}
\label{app:repro}

\paragraph{Base models.} Qwen3-8B (PoC / Phase-7) and Qwen3.5-9B (G2 / cross-dataset), in 4-bit via Unsloth.

\paragraph{LoRA.} Rank $r=8$, $\alpha=16$, dropout 0, applied to all attention and MLP modules ($\sim$0.42\%
of parameters trainable); per-segment (lifestage / NielsenIQ persona) for coverage, per-user only where
$\ge$300 pairs are available.

\paragraph{RAG.} A 1{,}024-dimensional embedding model into ChromaDB under cosine similarity; top-3 SKUs per
decision; present in \emph{both} training and inference.

\paragraph{Training.} SFTTrainer (TRL) with Unsloth; GRPO via TRL GRPOTrainer with Unsloth PatchFastRL, using
hand-written verifiable rewards (exact YES/NO and format; NO-rows weighted $\times 2$; Jaccard on B3),
$\sim$300--500 steps with 8 rollouts; assistant-only loss masking; \texttt{<think>} empty at training and
\texttt{enable\_thinking=False} at inference for fixed templates.

\paragraph{Evaluation.} vLLM (tensor-parallel 8 on H100$\times$8) batched at 2{,}000 prompts;
LLM-as-judge via DeepSeek-V3/V4 on BytePlus (rule-based where possible, LLM otherwise; 0.08\% judge-error
rate). Metrics: accuracy/precision/recall/F1 (purchase), hit-rate/Jaccard (basket), and AUC
(voucher; first-token log-probability of P(YES) plus a verbalized ``Likelihood: NN'' self-consistency).

\paragraph{Hardware.} Training and evaluation on RunPod H100$\times$8 and a local H20 96GB; a CUDA\_HOME fix
is required for flashinfer JIT, and batch inference avoids EngineCore OOM.